# Exploring Large Language Models in Healthcare: Insights into Corpora Sources, Customization Strategies, and Evaluation Metrics

**Shuqi Yang[1], Mingrui Jin[1], Shuai Wang[2], Jiaxin Kou[1], Manfei Shi[1], Weijie Xing[1],Yan Hu[1], Zheng Zhu*[1]**

1 School of Nursing, Fudan University, Shanghai, China
2 School of Nursing, Dali University, Yunnan, China
* Corresponding author

## Abstract

**Introduction:** Large Language Models (LLMs) are playing an increasingly vital role in healthcare, offering significant potential to transform various aspects of medical practice, particularly in question-answering systems. This study aims to review the corpora sources, customization techniques, and evaluation metrics of LLMs in healthcare.

**Methods:** A systematic search was conducted in PubMed/MEDLINE, Embase (OVID), Scopus (Elsevier), and Web of Science databases to identify studies between 2021 and 2024 that applied LLMs for delivering medical information. Data were extracted on LLMs' training corpora sources, construction methods, base models, AI customization techniques, and performance evaluation metrics.

**Results:** A total of 61 articles were included. The corpus sources used in the reviewed LLMs were categorized into four main types: real-world clinical resources (n=24), literature materials (n=34), open-source datasets (n=33), and web-crawled data (n=11). Notably, 44 studies employed a combination of multiple data sources to implement a more comprehensive model training approach. The primary techniques for constructing LLMs included pre-training, prompt engineering, retrieval-augmented generation, model fine-tuning, in-context learning, and offline learning. Fourteen studies used a single customization technique, while 41 studies combined these methods during model development. The evaluation metrics were classified into three main domains: 1) process metrics, 2) usability metrics, and 3) outcome metrics. The outcome metrics could also be divided into two categories: model-based outcomes and expert-assessed outcomes.

**Conclusion:** We identified critical gaps in corpus fairness, contributing to biases from geographic, cultural, and socio-economic factors. The reliance on unverified or unstructured data highlights the need for better integration of evidence-based clinical guidelines. Future research should focus on developing a tiered corpus architecture with vetted sources and dynamic weighting, while ensuring model transparency. Additionally, the lack of standardized evaluation frameworks for domain-specific models calls for comprehensive validation of LLMs in real-world healthcare settings.

## 1.Background

Large Language Models (LLMs) have become increasingly important in healthcare, offering substantial promise in revolutionizing various aspects of medical practice, particularly in the context of question-answering systems. These models, powered by deep learning algorithms and large-scale corpora of textual data, are capable of understanding and generating human-like text in response to queries. In healthcare, LLMs are being applied in a wide range of domains, such as clinical decision support, medical literature review, patient communication, and personalized treatment recommendations[1,2,3,4]. The ability of LLMs to process vast amounts of medical information and generate contextually relevant responses makes them invaluable tools for improving healthcare efficiency and patient outcomes.

However, a critical issue limiting the widespread use of LLMs in question-answering systems within clinical settings is the phenomenon of model hallucination. Hallucinations occur when the LLMs generates information that is factually incorrect, irrelevant, or fabricated, which can be particularly dangerous in the medical field[5]. For example, incorrect drug interactions or fabricated medical advice can lead to dire consequences for patients. Studies have highlighted the frequency of hallucinations in medical applications, with reports indicating that LLMs can produce erroneous responses up to 20-30% of the time in certain clinical tasks[1,2,3,4,5,6,7]. This presents a significant challenge for integrating LLMs into clinical workflows, where accuracy and reliability are paramount.

To mitigate these risks, significant improvements have been made in both the data sources used for training and the data enhancement methods. Research suggests that the quality and diversity of the training datasets are crucial for developing more accurate and context-specific models[8]. Curated datasets, such as those derived from clinical guidelines, patient records, and peer-reviewed medical literature, have been found to significantly improve model performance[9]. Additionally, recent advancements in retrieval-augmented generation (RAG) methods—where LLMs are augmented with external data sources in real time—have shown promise in improving the factual accuracy of model outputs. Techniques such as retrieval-based fine-tuning and knowledge-based augmentation help reduce the risk of hallucinations by providing the LLMs with access to reliable, up-to-date information[10]. These methods aim to bridge the gap between the model's internal knowledge and real-world clinical data, thus enhancing model accuracy and clinical relevance.

In addition, the evaluation of LLMs in healthcare remains an area of significant uncertainty. Current evaluation frameworks vary widely, with some focusing primarily on task-specific metrics such as accuracy and F1 score, while others emphasize clinical outcomes or user satisfaction[11]. There are some evaluation frameworks that heavily relying on task-specific metrics and expert assessments, while overlooking clinical relevance and ethic. Without a robust set of evaluation criteria, it is challenging to objectively compare the performance of different models and ensure that they meet the high standards required for clinical deployment.

Despite these advancements, there is currently no comprehensive review that systematically evaluates the existing corpora, data enhancement methods, and evaluation metrics for LLM-based question-answering systems in clinical settings. This gap in the literature underscores the need for a scoping review to synthesize current approaches, identify best practices, and propose directions for future research aimed at optimizing the application of LLMs in clinical practice. By consolidating existing evidence, this review will provide a clearer understanding of how to enhance the reliability and applicability of LLMs in real-world healthcare settings.

## 2. Methods

### 2.1 Study design

This scoping review was conducted in accordance with the JBI scoping review methodology[12] and followed to the Preferred Reporting Items for Systematic Reviews and Meta-Analyses extension for Scoping Reviews (PRISMA-ScR) checklist[13] for itsexecution and reporting (Appendix I).

### 2.2 Literature search

A three-step search strategy was employed to identify relevant literature. First, a preliminary search in PubMed/MEDLINE was conducted to identify a small set of relevant studies, which informed the development of a more comprehensive search strategy. Second, a comprehensive search across multiple databases, including PubMed/MEDLINE, Embase (Ovid), Web of Science, and Scopus, focusing on publications from 2021 to 2025. The starting point of 2021 was chosen as it marks a pivotal moment in the evolution of LLMs. In 2020, OpenAI released GPT-3, a model featuring 175 billion parameters, setting a new standard in the field. Subsequently, in 2021, Google introduced the Switch Transformer, which surpassed the trillion-parameter threshold for the first time. This milestone initiated a new era of exponential growth in the scale and capabilities of language models, driving significant advancements in AI research and applications, particularly in healthcare. Publications after 2021 are expected to capture the state-of-the-art developments and applications of these advanced models. Consequently, studies published before 2021 were excluded from our search, as they predominantly focused on earlier pre-trained models with smaller parameter sizes, which are less representative of the capabilities of contemporary LLMs.

In PubMed, a combination of free words and MeSH terms was utilized. The search terms included (("large language model"[Title/Abstract] OR "LLM"[Title/Abstract] OR "generative pre -trained transformer"[Title/Abstract]) OR ("GPT"[Title/Abstract] OR "ChatGPT"[Title/Abstract] OR "GPT-3"[Title/Abstract] OR "GPT-4"[Title/Abstract] OR "LLAMA"[Title/Abstract] OR "BERT"[Title/Abstract]

OR "Claude"[Title/Abstract])) AND (("Health"[MeSH Terms] OR "mental health"[Title/Abstract] OR "physical health"[Title/Abstract]) OR ("medical"[Title/Abstract] OR "medicine"[Title/Abstract] OR "healthcare*"[Title/Abstract])). Additionally, preprint repositories such as arXiv, SSRN, and Research Square were included in the search. A manual search was conducted on the reference lists of all included papers to identify any additional eligible studies. The complete search strategy is presented in Appendix II.

### 2.3 Inclusion and exclusion criteria

The inclusion criteria for this review were as follows: 1) studies with significant relevance to the development or application of LLMs; 2) research specifically related to the medical or healthcare field; 3) LLMs designed to provide medical or care-related information to users via dialogue interfaces; 4) articles written in English. The exclusion criteria were: 1) studies focused on early pre-trained models, such as GPT-2, which do not reflect the current advancements in LLMs' capabilities, and 2) non-primary research, including reviews, commentaries, or editorial pieces; 3) studies that did not specify the methods used for training LLMs; 4) studies that involved tasks such as using LLMs to build prediction models, classification models, or medical named entity recognition models.

### 2.4 Study selection

The search results were imported into EndNote (Clarivate Analytics, Philadelphia, PA, USA), where duplicate entries were removed. The study selection process was conducted independently by two reviewers (MJ and SW). Discrepancies between the reviewers were resolved through discussion, with any unresolved conflicts were adjudicated by a third reviewer (SY). The screening process began with an initial review of titles and abstracts, followed by a full-text assessment of studies deemed potentially relevant.

### 2.5 Data extraction

Three reviewers (SY, MJ and SW) used a customized data extraction form to collect relevant information. The extracted data included study titles, authors, citations, year of publication, country, research field, setting, study design, target population, LLMs corpus sources, construction methods, base models, data enhancement technologies, quantitative evaluation indicators, and qualitative evaluation indicators. Data extraction was independently performed by two reviewers, with discrepancies resolved through discussion. In cases where a consensus could not be reached, a third reviewer (ZZ) was consulted to make the final decision.

## 3. Result

### 3.1 Literature search

A total of 14,744 studies were identified through the database search (Figure 1). After duplicates were removed and titles and abstracts were screened, 248 studies were selected for full-text review. Of these, 61 articles met the inclusion criteria and were included in the analysis. The excluded studies, along with the reasons for their exclusion, are detailed in Appendix III .

3.2 Study description

Table 1 provides an overview of the general characteristics of the studies included in the final analysis. The number of studies on the application of LLMs in healthcare steadily increased showed a steady increase over the years, with one study published in 2022, 14 studies in 2023, and 46 studies in 2024. The studies were distributed across multiple countries, with the United States contributing the highest number (N=26), followed by China (N=18). Other countries included Australia (N=3), Korea (N=3), Canada (N=3), Italy (N=2), and Germany (N=2).

3.3 Application areas and target populations of LLMs in healthcare

The characteristics of the included studies are summarized in Table 1. Among the 61 studies, LLMs were developed in three primary types: single-domain applications, multi-domain applications, and unspecified domain application. Single-domain applications focused on specific medical fields. Oncology was the most frequently studied domain, represented by seven studies[17,41,46,47,55,60], followed by ophthalmology with six studies[34,37,40,43,57,69], hepatology[24,54]and orthopedics[16,66]were each addressed in two studies, while the following fields were each represented by a single study: neurosurgery[20], otolaryngology[28], endocrinology[29], nephrology[30], pediatrics[48], and neurology[49]. Multi-domain applications included a broader range of disciplines, including diverse diseases, public health, traditional Chinese medicine (TCM), mental health, and medical education. Four studies specifically addressed mental health issues, with Kharitonova's study focusing on depression and Attention-Deficit/Hyperactivity Disorder (ADHD) [25,39,50,58]. Five studies were dedicated to TCM[22,35,53,56,64], and three studies explored medical education, two of which specifically examined specialized training in anatomical sciences and surgical and anesthetic education[18,21,44]. Other domains included radiology (n=3)[15,33,62], pharmaceutical sciences (n=2)[31,42]and nutriology (n=2)[14,71]. Additionally, 16 studies focus on unspecified domain applications, which are widely utilized across various medical fields to address diverse challenges and enhance clinical practices[19,23,26,27,32,36,38,45,51,58,60,62,64,66,71,73.]

Sixty-one studies that developed LLMs primarily targeted two groups: patients and individuals seeking medical support, and healthcare providers. The majority of the LLMs (33 studies) were designed to assist healthcare providers in efficiently obtaining and analyzing medical information[18,22,24,26,28,32,33,34,35,38,42,44,47,51,54,55,57,59,62,66,67,70,72-74]. Ten studies focused on LLMs designed to assist patients or individuals seeking medical support, enabling them to access information in a more accessible and comprehensible manner. Eighteen studies constructed LLMs that served to both groups. These LLMs typically lacked a defined target audience and instead focused on providing relevant knowledge tailored to the specific needs of their respective fields.

## 3.3 Corpus sources in LLMs research

Table 1 also summarized the corpus sources of the included LLMs, categorized into four primary types: real-world clinical resources, literature materials, open-source datasets, and web-crawled data.

Real-world clinical resources constituted a key category, with 24 studies[14, 17,24,27,28,30,32,33,38,39,41,44,48,55,56,67,69,71,74] utilizing clinical data such as Electronic Health Records (EHRs), diagnostic reports, real-world case data, medical podcast, and expert-drafted clinical cases from hospitals worldwide.

Literature materials were another key sources of healthcare LLMs' training, with 34 studies[16, 18,21,23,24,26,28-31,33,34,36,38,40,41,47-49,52-54,56-60,63,66,69,72] leveraging verified, high-level, evidence-based literature. These sources included textbooks, clinical guidelines, peer-reviewed journal articles, and other authoritative resources.

Open-source datasets were widely used in research, with 33 studies[15, 18, 19,27,28,31,32,34,36,38-40,42,44,45,48-50,52,53,56,57,59-64,69,70] leveraging publicly available datasets. Datasets from 24 studies included validated question-answer (QA) pairs derived from established medical QA datasets, such as SQuAD-IT, MedMCQA, MedQA-USMLE, PubMedQA, emrQA, and other specialized-domain QA benchmarks[15, 19,22,27,28,31,32,34,36,38,40,44,45,48,49,53,56,57,59,61,63,70]. Additionally, ten studies[17,45,47,57,64,65,67,71,73,74] used virtual synthetic datasets, including virtual doctor-patient dialogue collections, virtual cases, and medical information generated and organized through simulated data. Two studies[45,57] combined real-world and synthetic datasets to optimize training.

Web-crawled data was another important source, utilized in eleven studies[18,20,21,25,26,27,43,51,57,61,65]. This data was scraped from publicly available medical content on the internet, including health forums, medical websites, and social media platforms. Two studies[21,25] incorporated unverified data, while six[37,43,46,51,57,61,65] focused on unstructured datasets, such as real-world doctor-patient dialogue datasets.

Among the studies, 44 combined multiple types of data sources to adopt a more comprehensive model training approach. Six studies relied only on academic literature sources[23,29,54,58,66,72], four used only open-source datasets[22,32,62,70], three used real-world clinical resources exclusively[14,47,55], three relied solely on web-crawled data[25,43,51], and one study utilized only virtual datasets[73].

## 3.4 Base model of LLMs

Table 2 showed the base models of the included LLMs. These models could be grouped into several key categories. The GPT series, including GPT-3, GPT-3.5, and GPT-4, accounted for 26 studies[16-18,20,23,24,28-34,39,41,42,49,50,54,58,60,71-74]. The LLaMA series, including 15 models such as

LLaMA-2 and its variants (7B, 13B, and 70B) [14,27,36,44,51,55,58,61,64-68]. Additionally, some architectures were developed based on the LLaMA framework. The Mistral series, with four models, focused on instruction tuning to enhance the models ' ability to follow user commands[19,44,47,59]. The Baichuan series, consisting of five models, emerged as a prominent Chinese domestic base mode[22,35,48,53,69]. The Vicuna[21,70], Ziya-LLaMA-13B[56,63] and ChatGLM[37,40] series were each used in two studies. Each of the following smaller models was used in a single study: IT5[15], Aquila-7B[45], Alpaca-7B[62], RoBERTa[43], BERT[38], Gemini Pro[47], WenZhong[25], and PanGu pre-trained models[25].

## 3.5 AI customization techniques of LLMs in healthcare

The AI customization techniques of LLMs are summarized in Table 2. The primary techniques for constructing LLMs included pre-training LLMs, prompt engineering, RAG, model fine-tuning, in-context learning, and offline learning. Fourteen studies independently used one of the AI customization techniques[18,21,34,40-42,51,54,55,64-66,71,72], while the other 41 studies combined these methods during the model development process[14-17, 19,20,22-33,35-39,43-50,52,53,55-63,67-70,73,74].

### 3.5.1 Pre-training LLMs

Sixteen studies focused on model pre-training[15, 19,22,25,35,38,44-46,48,53,56,57,63,68], which involves unsupervised pre-training on unlabeled data, enabling the models to learn general features before fine-tuning for specific tasks.

### 3.5.2 Prompt engineering

A total of 37 studies utilized prompt engineering, with one study[71] employing this technique independently, while the others combined it with other methods[14-20,22-33,36,39,40,43,47,49,50,52,53,58-61,63,67,68,73,74]. Zero-shot prompting was the most commonly used approach, applied in eleven studies[14-17, 19,20,63,71]. Seven studies [17,23,24,29,59,63,73] employed few-shot prompting. Five studies used Chain-of-Thought (CoT) prompting to guide the model through a structured reasoning process[22,30,47,49,71]. Eight studies[26,28,33,59,61,67,74] used instruction-based prompting, which involved providing clear, task-specific instructions to guide the model's responses. In contrast, template-based prompting, used in five studies[43,50,52,53,58], relied on a predefined template to structure the prompt consistently. For soft tuning, prompt tuning was employed in four studies[32,40,68], enabled efficient task adaptation through gradient-based updates, optimizing memory and storage usage.

### 3.5.3 RAG

Twenty studies employed RAG[14, 17,23,24,28-31,34,37,42,47,49,54,58,60,69,70,72]. Twelve studies using naive RAG adopted a basic retrieval method, where knowledge from external sources was encoded as vector representations using sentence transformers[19,27,29,31-34,50,58,62,73,74]. Two studies[23,49] employed Retrieval-Augmented Generation with Knowledge Graphs (RAG-KG), which incorporated structured data and semantic relationships to establish disease-related logic and improve retrieval accuracy. One study focused on RAG combined with a multi-agent orchestration system, which allows for more sophisticated and efficient task execution[42].

Additionally, five studies highlighted the unique capabilities of RAG when integrated with LangChain technology, which enhanced data retrieval and integration from external tools such as Application Programming Interfaces (APIs) and databases[14,34,37,60,72].

3.5.4 Model fine-tuning

Thirty-seven studies used model fine-tuning technique[14,15,19,21,22,25,27,32,35-39,41,43-46,48,50-53,55-57,59,61-69,74]. Model fine-tuning in included studies typically involves two main types: Supervised Fine-Tuning (SFT) and Reinforcement Learning from Human Feedback (RLHF).

SFT, used in 13 studies, adapted LLMs to healthcare-specific tasks by training them on instruction-response pairs[15,27,45,48,50,53,56,59,63,65,69,74]. Eight studies used instruction tuning, particularly full-parameter fine-tuning, to further refine the model, enabling them to generate accurate and contextually relevant responses based on medical instructions[14,36,46,50,53,62,64,68].

To address the computational constraints of LLMs, 18 studies used Parameter-Efficient Fine-Tuning (PEFT) methods[21,22,37,43,48,50-52,55-57,59,62,65,67,69]. Among them, 14 studies used Low-Rank Adaptation (LoRA), reducing computational costs with low-rank matrices[21,22,27,37,43,46,48,50,51,55,56,62,67,69]. Additionally, 5 studies used Quantized Low-Rank Adaptation (QLoRA), incorporating quantized parameters to optimize fine-tuning[22,52,57,59,65].

RLHF, used in 4 studies[45,56,63,65], refines models by aligning them with human preferences and medical standards. One study also employed Direct Preference Optimization (DPO) to further enhance performance based on human evaluations[45]. Additionally, a study applied Reinforcement Learning from AI Feedback (RLAIF) after SFT, optimizing the model to generate patient-friendly and doctor-like responses with professional medical knowledge and diagnostics[74].

3.5.5 In-context learning

In-context learning in healthcare LLMs was employed in four studies to improve model performance by utilizing relevant external context in real-time inputs[20,33,70,73]. One study combined in-context learning with RAG to incorporate contextually relevant data for accurate responses[66]. Additionally, two studies integrated in-context learning with Langchain framework, enhancing data retrieval from external tools[20,73].

3.5.6 Offline learning

One study applied offline knowledge distillation to enhance user intent extraction by transferring knowledge from a teacher model, improving accuracy without real-time updates, and supporting tasks like PRO data collection within computational limits[17].

3.6 Evaluation approaches of LLMs in healthcare

The evaluation approaches of LLMs in healthcare included three approaches. Human-based evaluation focusing on task examinations was used in 36 studies [14-20,22,24-29,31-34,36,37,39-41,45,48,50,53,59,62,63,64,67-69,73,74], while 15 studies used human-based evaluation for case analysis[19,22,25,30,35,36,42,46,50,53,54,66,69-71]. In addition, 43 studies employed model-based evaluation for the same purpose[15, 16, 18-22,24,25,27,29,32,35,38,39,41-45,48-53,55-58,60,61,63,65,67-72,74]. Eleven studies combined multiple evaluation methods to assess model performance[19,22,25,35,36,42,50,53,69-71].

The expert-assessed outcome metrics evaluated using five primary measurement methods. Likert scales or binary/multiple-choice evaluation scales were employed in 18 studies[15, 17, 19,20,25,26,27,31,32,37,42,50,59,62,63,64,69,73]. Unstructured feedback was collected

in 16 studies, offering more detailed and subjective insights into model performance[19,22,25,30,35,36,39,42,47,53,54,66,70,71]. The comparison of model responses with human expert-provided answers served as the gold standard in 15 studies, offering a direct benchmark for model performance[14-17,21,22,24,25,28,29,33,34,36,67,68]. Performance rankings based on comparisons between different models were used in two studies[36,48]. The comparison of predicted versus actual responses in two studies aimed to evaluate the models' accuracy and reliability in forecasting outcomes[25,45]. Additionally, the presence or absence of specific categories within model responses was assessed in four studies[14, 16, 18,47].

### 3.7 Evaluation metrics of LLMs in healthcare

The reviewed studies predominantly employed systematic evaluation approaches, with only one study excluding model performance assessment[23]. Based on our analysis, the evaluation metrics utilized across these studies can be categorized into three primary domains: 1) process metrics, 2) usability metrics, and 3) usability metrics.

#### 3.7.1 Process metrics

As systematically categorized in Table 3, the process evaluation metrics for LLMs in healthcare applications included ten distinct dimensions. The most frequently used process evaluation metrics were repetition testing turns[33,34,45,53,56,63,74] and model size [32,38,44,46,53,68,69], each being evaluated in seven independent studies (n=7). Repetition testing turns were primarily employed to assess the model's consistency maintenance during iterative clinical interactions, while model size evaluations focused on computational capacity analysis and resource allocation requirements. Secondary process metrics included training parameters[38,46,56,57,58], which were examined in five studies (n=5) to investigate the correlation between specific configurations and model performance. Response time metrics[17,20,21,33,46], evaluated in four studies (n=4), provided critical insights into the models' operational efficiency in time-sensitive clinical scenarios. Additionally, output length analysis[21,39,65], conducted in three studies (n=3), served as a quantitative measure for evaluating response conciseness and clinical relevance.

3.7.2 Usability evaluation metrics

The usability evaluation metrics of LLMs in healthcare are summarized in Table 3. The most commonly used metric was user helpfulness, assessed in nine studies [16,25,27,39,48,50,62,64,73]. Other metrics, such as user intent fulfillment[16], response costs[33], personalization[71], and interactivity[71], were each evaluated in one study.

3.7.3 Outcome metrics

Outcome metrics were categorized into two main types: model-based outcomes and expert-assessed outcomes. As shown in Table 4, the model-based outcome metrics for LLMs in healthcare were further classified into two groups: (1) general domain standard metrics and (2) automated metrics evaluation using structured prompts in LLMs.

We identified 17 model-based outcomes across the included studies. Accuracy emerged as the most prevalent metric (n=20)[15, 16, 18, 19,20,21,29,32,38,44,49,50,52,57,58,63,68,70,71,74]. Precision (n= 13)[17,22,27,29,32,38,43,46,51,57,61,67,72], recall (n= 13) [17,22,27,29,32,38,43,46,51,57,61,67,72], and F1 score (n=15)[15, 17,22,27,29,32,38,42,43,46,51,57,61,67,72] were typically used together as components of the Bidirectional Encoder Representations from Transformers (BERT) score evaluation system. Ten studies employed Bilingual Evaluation Understudy (BLEU)[24,35,39,43,46,48,50,63,69,74], and eleven studies employed Recall-Oriented Understudy for Gisting Evaluation (ROUGE)[24,25,35,43,46,48,55,63,69,72,74]. Additionally, four studies employed distinct[25,43,48,74] and three studies used Metric for Evaluation of Translation with Explicit Ordering (METEOR) score[24,50,72] as supplementary metrics to assess diversity and semantic accuracy. Addtionally, three studies used cosine similarity to evaluate semantic alignment between model-generated responses and reference answers in healthcare-specific tasks[47,50,65].

The automated evaluation approach utilizing structured prompts involved employing LLMs to assess model-generated responses. A notable application of this method was the comparative evaluation of PediatricsGPT-13B against baseline models through GPT-4-based assessment of dialogue responses[48]. The analysis revealed the following distribution of evaluation metrics: Fluency/smoothness (n=5)[45,48,53,56,72], relevance (n=3)[45,53,72] and accuracy (n=2)[48,53], completeness (n=2)[45,53], consistency (n=2)[48,72] and proficiency (n=2)[45,53]. Safety[56], professionalism[56], and coherence[72] were each used in a single study.

Table 5 presents the expert-assessed outcome metrics for LLMs in healthcare, categorized into five domains. The Information quality domain included five sub-metrics: accuracy (34 studies)[13-17, 19-22,24-29,31,33-37,40-42,48,53-54,59,63,66-69,73,74], completeness (n=7)[15, 17,20,26,37,41,63], relevance (n=12)[16,20,25,26,31,32,37,41,50,62,69,70], comprehension (n=9)[16, 17,29,36,59,62,66,69,71], and consistency (n=8)[17, 19,25,28,32-34,47]. These metrics evaluated the model's capacity to deliver accurate, comprehensive, and contextually appropriate information within healthcare settings. Safety and risk domain focused on four critical aspects: safety (n=4)[17,28,48,64], risk (n=7)[16, 19,29,31,35,39,69], bias (n=6)[16, 19,29,50,63,69], and hallucination (n=5)[14, 16,24,31,47]. This domain evaluated potential hazards, including output inaccuracies, inherent biases, and other risks associated with LLM. The reasoning and justification domain comprised two key components: reasoning (n=6)[16,25,29,35,42,69] and provision of rationales with citations (n=6)[18,26,31,36,47,69]. These metrics examined the model's ability to provide logically sound explanations supported by credible references for its outputs. Communication quality included five sub-metrics: quality (n=9)[22,25,28,30,35,39,45,48,53], empathy (n=4)[19,27,35,50], readability (n=14)[20,22,25,31,32,36,37,41,50,59,62,63,64,73], responsiveness (n=4)[25,27,35,37], and creativity (n=1)[22]. These metrics evaluated the model's effectiveness in delivering information, considering aspects such as response tone, clarity, and adaptability to user needs. Lastly, the cultural fidelity (n=1)[22] assessed the LLM's capability to generate culturally appropriate and sensitive responses within healthcare contexts.

## 4. Discussion

This is the first scoping review to explore the corpus sources, AI customization techniques, and evaluation metrics of healthcare LLMs. We identified four main types of corpus sources: real-world clinical resources, literature materials, open-source datasets, and web-crawled data. Regarding AI customization techniques, the studies highlighted several methods used in constructing these models, including pre-training, prompt engineering, RAG, fine-tuning, in-context learning, and offline learning. In terms of performance evaluation, the evaluation metrics applied were grouped into three main categories: process metrics, outcome metrics, and usability metrics.

We found that 72% of the reviewed LLMs relied on multiple data sources, with literature materials and open-source datasets being the most frequently utilized. While integrating diverse data sources can enhance model performance and applicability, it also raises significant equity concerns, as it inherently embeds biases related to geographic, cultural, and socio-economic factors. Despite incorporating various types of data, the majority of these sources are concentrated in high-income countries or specific regions, leading to potential biases. For instance, previous analyses of the PubMed dataset have highlighted substantial gaps in socio-demographic representation, including nationality, gender, and geographic region, within these datasets[75]. Such biases[76] disproportionately affect underrepresented populations, particularly those in low- and middle-income countries, where clinical features and healthcare challenges may be systematically overlooked. The prerequisite for data openness is the establishment of digital infrastructure. Zharima et al.'s study[77] showed that even in South Africa, where policies to promote healthcare digitization were widely formulated, the adoption rate of Electronic Health Record (HER) systems remained low, and progress in building a digital ecosystem was slow. This lack of representation perpetuates healthcare inequities, as LLMs trained on narrow datasets are ill-equipped to address the unique needs of diverse patient groups. Consequently, reliance on data from certain regions or specific sources exacerbates disparities in healthcare resource allocation, reinforcing systemic biases and limiting the potential of AI-driven solutions to achieve equitable outcomes. This narrow focus perpetuates healthcare inequities by failing to address the distinct challenges faced by underrepresented patient groups, thereby hindering the broader applicability of AI-driven healthcare solutions. To address these challenges, it is imperative to integrate a more comprehensive array of data sources from varied geographic regions and socio-economic backgrounds. Only by prioritizing diversity and inclusivity in dataset construction can we ensure that LLMs are capable of serving global populations equitably and mitigating the entrenched biases that currently hinder their broader applicability.

In addition, we found that, aside from the six studies relying solely on academic literature sources[23,29,54,58,66,72], the remaining 90% of studies incorporated at least one type of unverified or unstructured data. The quality and accuracy of LLMs are inherently tied to the quality of their training data. Unverified or unstructured data can significantly compromise model performance, leading to unreliable or even harmful outputs. Brown et al.'s [78]study showed the importance of filtering out low-quality information to enhance model performance. However, many corpus sources currently used to train healthcare-specific LLMs lack professional medical review. For instance, web-crawled content, such as medical forums or doctor-patient dialogues, often includes clinically unvalidated recommendations[79]. Furthermore, unstructured data, which lacks a standardized format, forces the model to rely heavily on contextual interpretation during processing, increasing the risk of ambiguity and misinterpretation. For example, the abbreviation "CP" in clinical notes could represent either "chest pain" or "cerebral palsy", while non -standardized entries like "BID" for "twice daily" could lead to dosage errors. Without sufficient context, LLMs may fail to accurately interpret such terms, potentially resulting in clinically significant errors. these correctly. While frameworks like Wiest et al.'s have improved the accessibility and utility of unstructured medical text data through LLMs, their effectiveness still depends on high-quality input data. Similarly, synthetic datasets, such as virtual doctor-patient dialogues, often oversimplify the complexity of real-world medical conditions, further limiting their utility. Alber's[80] recent study found that introducing even a minimal amount of medical misinformation—as little as 0.001% of training tokens—can result in models that propagate harmful errors. When LLMs are trained on erroneous or unverified data, they struggle to differentiate between evidence-based facts and unsubstantiated anecdotes, often presenting inaccurate recommendations with unwarranted confidence.

To mitigate these risks, it is important to integrate evidence-based principles into the development of healthcare-specific LLMs. Incorporating clinical practice guidelines ensures that model outputs align with the latest clinical standards while also addressing the nuanced interplay between patient preferences, values, and evidence-based care[20]. Future research should prioritize the implementation of a tiered corpus architecture, where core training data is derived from rigorously vetted sources and supplemented by carefully curated unstructured data. This approach would strike a balance between model adaptability and safety. For instance, training data could be organized into tiers based on reliability (e.g., Level I: evidence-based guidelines and expert consensus; Level II: electronic health records; Level III: curated forum data), with dynamic weighting applied to each tier. Additionally, tracking the provenance of information throughout model development and transparently presenting the reasoning chain behind outputs would ensure that conclusions are grounded in robust, evidence-based sources.

Our study highlights that most LLMs evaluation metrics focus on a single dimension, such as repetition testing turn, accuracy or helpfulness, which is insufficient given the complexity of healthcare. A multidimensional approach is necessary to meet the rigorous demands of clinical practice. We identified three critical metric categories in current evaluation frameworks for healthcare LLMs. For process metrics, common indicators like model size measure efficiency and stability but do not assess real-world clinical effectiveness[32,38,69,74]. For outcome metrics, model-based metrics predominantly rely on general domain standards, such as accuracy and F1-score, as well as automated evaluations via structured prompts like BLEU and ROUGE[15,24,48,74]. However, these metrics fail to address clinical semantic accuracy, as evidenced by the absence of healthcare-specific measures, such as clinical guideline compliance and differential diagnosis validity. Expert assessments of LLMs, covering information quality, safety, reasoning, communication, and cultural fidelity, are crucial but can be subjective and biased. Mitigating subjectivity is essential for ensuring reliable model performance.

Moreover, hallucinations, a major issue in LLMs, are inadequately addressed by current metrics. While traditional metrics measure content correctness, they miss hallucinations that could endanger patient safety[14]. Future research must develop methods for detecting and mitigating hallucinations, integrating these into training and evaluation stages. Improving transparency and interpretability is key to reducing hallucinations, ensuring reliable, safe clinical decision-making, and supporting reasoning with high-quality evidence[18,25,31,35,47]. Finally, usability metrics focus on user experience but often overlook critical aspects like supporting clinical decisions and adapting to diverse healthcare environments. This narrow metrics needs expansion to capture full clinical applicability.

Previous research has proposed frameworks for evaluating LLM construction. For instance, Choi et al. [81]utilized the SERVQUAL (Service Quality) framework to assess ChatGPT's service quality, while Long et al. applied the CVSC (Concordance, Validity, Safety, and Concordance) framework to evaluate their models[28]. Tam et al. [82]developed the QUEST (Quality Evaluation for Service Training) framework for human evaluation of LLMs. However, no universally applicable framework exists for the systematic evaluation of LLMs. Furthermore, current evaluation methods are inefficient, relying heavily on static datasets, supervision signals, or manual expert assessments. Some studies have used LLMs to evaluate responses from constructed models, performing deep interactions between domain-specific models and LLMs through single or multiple instructions, followed by performance evaluation[83,84].A previous review noted that LLMs tend to rate chatbot responses higher than human gold standards, suggesting the need for further scrutiny of LLMs as an evaluation tool. Future evaluation frameworks should integrate multidimensional metrics that assess both technical performance and clinical applicability. Additionally, a hybrid evaluation approach combining automation and expert input could reduce manual workload while ensuring professionalism. Finally, ethical considerations and fairness must be incorporated to account for diverse populations and contexts.

## Limitations

This scoping review is the first comprehensive exploration of corpus sources, customization techniques, and evaluation metrics for healthcare LLMs. However, our review has several limitations. First, we did not explore the evaluation requirements for specific healthcare tasks, such as clinical decision support and medical record summarization. Different tasks may require distinct evaluation criteria to ensure accuracy, reliability, and clinical applicability. Future research should refine evaluation frameworks to incorporate both overarching principles and task-specific metrics, thereby enhancing their relevance to real-world healthcare applications. Additionally, given the rapid evolution of LLMs' architectures, our analysis, based on methodologies up to 2024, may not fully capture the latest developments. This may lead to an incomplete understanding of the healthcare LLMs' landscape. Future research should adapt to emerging frameworks and techniques to ensure accurate evaluation and clinical applicability.

## Conclusion

This scoping review analyzed 61 studies on corpus sources, customization techniques, and evaluation metrics for LLMs in healthcare. A significant gap was found in the fairness of corpus usage, leading to biases tied to geographic, cultural, and socio-economic factors. Moreover, most studies incorporated unverified or unstructured data, highlighting the need for stronger integration of evidence-based sources, especially high-level clinical guidelines. Future research should focus on developing a tiered corpus architecture with integrates rigorously vetted sources and dynamic weighting, ensuring transparency in model reasoning and information provenance. Additionally, the lack of standardized evaluation systems for vertical-specific models underscores the need for comprehensive frameworks and real-world validation of healthcare LLMs.

Table 1 Overview of characteristics of the included LLMs

| Author | Application areas | Target population | LLMs corpus sources | | | | |
|---|---|---|---|---|---|---|---|
| | | | **Real-world clinical resources** | **Literature sources** | **Virtual datasets** | **Open-source datasets** | **Web crawled data** |
| Alkhalaf, et al (2024, Australia)[14] | Nutriology | ① | √ | N/P | N/P | N/P | N/P |
| Bergomi, et al (2024, Italy)[15] | Radiology | ① | √ | N/P | N/P | √ | N/P |
| Chen, et al (2024, China)[16] | Sports medicine and orthopedics | ① | √ | √ | N/P | N/P | N/P |
| Chen, et al (2024, China)[17] | Oncology | ① | √ | N/P | √ | N/P | N/P |
| Collins, et al (2024, USA)[18] | Anatomical sciences education | ② | N/P | √ | N/P | √ | √ (unverified data) |
| Griot, et al (2024, Belgium)[19] | Unspecified domain | ② | N/P | √ | N/P | √ | N/P |
| Guo, et al (2023, Canada)[20] | Neurosurgery | ② | N/P | √ | N/P | N/P | √ |
| Guthrie, et al (2024, USA)[21] | Surgical and anesthetic education | ② | N/P | √ | N/P | N/P | √ |
| Hua, et al (2024, China)[22] | TCM | ② | N/P | N/P | √ | N/P | N/P |
| Kakalou,et al (2024, Greece)[23] | Unspecified domain | ① | N/P | √ | N/P | N/P | N/P |
| Kresevic,et al (2024, Italy, USA)[24] | Hepatology | ② | √ | √ | N/P | N/P | N/P |
| Lai, et al (2024, Australia)[25] | Mental health | ①② | N/P | N/P | N/P | N/P | √ (unverified data) |
| Li, et al (2024, USA)[26] | Unspecified domain | ② | N/P | √ | N/P | N/P | √ |
| Liu, et al (2024, USA)[27] | Unspecified domain | ①② | √ | N/P | N/P | √ | N/P |
| Long, et al (2024, Canada, USA)[28] | Otolaryngology | ② | √ | √ | N/P | √ | N/P |
| Mashatian,et al (2024, USA)[29] | Endocrinology | ① | N/P | √ | N/P | N/P | N/P |
| Miao,et al (2024, USA)[30] | Nephrology | ② | √ | √ | N/P | N/P | N/P |
| Murugan,et al (2024, USA)[31] | Pharmaceutical sciences | ①② | N/P | √ | N/P | √ | N/P |
| Peng, et al (2023, USA)[32] | Unspecified domain | ② | √ | N/P | N/P | √ | N/P |
| Rau, et al (2023, Germany)[33] | Radiology | ② | √ | √ | N/P | N/P | N/P |
| Singer, et al (2023, USA)[34] | Ophthalmology | ①② | N/P | √ | N/P | √ | N/P |
| Tan, et al (2024, China)[35] | TCM | ② | N/P | N/P | N/P | √ | N/P |
| Wu,et al (2024, China)[36] | Unspecified domain | ② | N/P | √ | N/P | √ | N/P |
| Xu, et al (2024, China)[37] | Ophthalmology | ① | N/P | √ | N/P | N/P | √ (doctor-patient dialogue datasets) |
| Yang, et al (2022, USA)[38] | Unspecified domain | ② | √ | √ | N/P | √ | N/P |
| Yu, et al (2024, UK)[39] | Mental health | ①② | √ | N/P | N/P | √ | N/P |
| Zheng, et al (2024, China)[40] | Ophthalmology | ①② | N/P | √ | N/P | √ | N/P |
| Zhu, et al (2024, USA)[41] | Oncology | ①② | √ | √ | N/P | N/P | N/P |
| Choi, et al (2024, USA)[42] | Pharmaceutical sciences | ② | N/P | N/P | N/P | √ | N/P |
| Fu, et al (2024, China)[43] | Ophthalmology | ①② | N/P | N/P | N/P | N/P | √ (doctor-patient dialogue datasets) |
| Jia, et al (2024, USA)[44] | Medical education | ② | √ | N/P | N/P | √ | N/P |
| Zhao, et al (2024, China)[45] | Unspecified domain | ② | √ | N/P | √ | √ | N/P |
| Li, et al (2024, USA)[46] | Oncology | ①② | √ | N/P | N/P | N/P | N/P |
| Lammert, et al (2024, Germany)[47] | Oncology | ② | √ | √ | N/P | √ | N/P |
| Yang, et al (2024, China)[48] | Pediatrics | ①② | √ | √ | √ | N/P | N/P |
| Li, et al (2024, USA)[49] | Neurology | ①② | N/P | √ | N/P | √ | N/P |
| Na, et al (2024, Australia)[50] | Mental health | ① | N/P | N/P | N/P | √ | N/P |
| Jia, et al (2024, China)[51] | Oncology | ② | N/P | N/P | N/P | N/P | √ (doctor-patient dialogue datasets) |
| Bhatti, et al (2023, Canada)[52] | Unspecified domain | ①② | N/P | √ | N/P | √ | N/P |
| Chen, et al (2023, China)[53] | TCM | ①② | N/P | √ | N/P | √ | N/P |
| Ge, et al (2023, USA)[54] | Hepatology | ② | N/P | √ | N/P | N/P | N/P |
| Liu, et al (2023, USA)[55] | Oncology | ② | √ | N/P | N/P | N/P | N/P |
| Yang, et al (2023, China)[56] | TCM | ①② | √ | √ | N/P | √ | √ (doctor-patient dialogue datasets) |
| Haghighi, et al (2024, USA)[57] | Ophthalmology | ② | N/P | √ | √ | √ | N/P |
| Kharitonova, et al (2024, Spain)[58] | Mental health (depression and ADHD) | ② | N/P | √ | N/P | N/P | N/P |
| Labrak, et al (2024, France)[59] | Unspecified domain | ② | N/P | √ | N/P | √ | N/P |
| Lee, et al (2024, Korea)[60] | Oncology | ① | N/P | √ | N/P | √ | N/P |
| Li, et al (2023, USA)[61] | Unspecified domain | ①② | N/P | N/P | N/P | √ | √ (doctor-patient dialogue datasets) |
| Liu, et al (2024, USA)[62] | Radiology | ② | N/P | N/P | N/P | √ | N/P |
| Tian, et al (2024, China)[63] | Unspecified domain | ①② | N/P | √ | N/P | √ | N/P |
| Wang, et al (2023, China)[64] | TCM | ①② | N/P | N/P | √ | √ | N/P |
| Wang, et al (2023, China)[65] | Unspecified domain | ①② | N/P | N/P | √ | N/P | √ (doctor-patient dialogue datasets) |
| Kim (2023, Korea)[66] | Orthopedics | ② | N/P | √ | N/P | N/P | N/P |
| Kumichev, et al (2024, Russia)[67] | Unspecified domain | ② | √ | N/P | √ | N/P | N/P |
| Kweon, et al (2024, Korea)[68] | Healthcare informatics | ② | N/P | N/P | √ | √ | N/P |
| Luo, et al (2024, China)[69] | Ophthalmology | ② | √ | √ | N/P | √ | N/P |
| Shi, et al (2024, USA)[70] | Clinical healthcare | ② | N/P | N/P | N/P | √ | N/P |
| Yang, et al (2024, USA)[71] | Nutriology | ① | √ | N/P | √ | N/P | N/P |
| Lozano, et al (2023, USA)[72] | Unspecified domain | ② | N/P | √ | N/P | N/P | N/P |
| Yu, et al (2024, USA)[73] | Clinical risk management | ② | N/P | N/P | √ | N/P | N/P |
| Zhang, et al (2023, China)[74] | Unspecified domain | ② | √ | N/P | √ | N/P | N/P |

① Patients and individuals seeking medical support; ② Healthcare providers Information not provided is marked by N/P.

TCM: Traditional Chinese Medicine; LLMs: Large Language Models; CKD: Chronic Kidney Disease

Table 2 Overview of Included LLMs: base model and AI customization techniques

| Author | Base model | AI customization techniques | | | | | |
|---|---|---|---|---|---|---|---|
| | | Pre-training LLM | Prompt engineering | RAG | Model fine-tuning | In-context learning | Offline learning |
| Alkhalaf, et al (2024, Australia)[14] | Llama 2 model | N/P | √ (zero-shot) | √ (LangChain technology) | √ (instruction tuning) | N/P | N/P |
| Bergomi, et al (2024, Italy)[15] | IT5 Base version model (220 M of parameters) | √ | √(zero-shot) | N/P | √ (SFT) | N/P | N/P |
| Chen, et al (2024, China)[16] | GPT-4-1106-preview model | N/P | √(zero-shot) | √ | N/P | N/P | N/P |
| Chen, et al (2024, China)[17] | ChatGPT | N/P | √ (zero-shot, few-shot) | N/P | N/P | N/P | √ (offline knowledge distillation) |
| Collins, et al (2024, USA)[18] | GPT-4 | N/P | √ (prompt tuning) | N/P | N/P | N/P | N/P |
| Griot, et al (2024, Belgium)[19] | Mistral LLaMA 7b | √ | √(zero-shot) | N/P | √ | N/P | N/P |
| Guo, et al (2023, Canada)[20] | GPT-3 | N/P | √ (zero-shot) | N/P | N/P | √ (LangChain technology) | N/P |
| Guthrie, et al (2024, USA)[21] | Vicuna v1.5 which is based on the LLaMA architecture | N/P | N/P | N/P | √ (LoRA) | N/P | N/P |
| Hua, et al (2024, China)[22] | Baichuan2-13B-Base | √ | √ (CoT technique) | N/P | √ (QLoRA) | N/P | N/P |
| Kakalou,et al (2024, Greece)[23] | GPT-4 | N/P | √(few-shot) | √ (KG) | N/P | N/P | N/P |
| Kresevic,et al (2024, Italy, USA)[24] | GPT-4 Turbo | N/P | √ (few-shot) | √ | N/P | N/P | N/P |
| Lai, et al (2024, Australia)[25] | WenZhong and PanGu pre-trained models | √ | N/P | N/P | √ | N/P | N/P |
| Li, et al (2024, USA)[26] | GPT-4 Turbo | N/P | √ (instruction-based prompting) | √ | N/P | N/P | N/P |
| Liu, et al (2024, USA)[27] | LLaMA-65B | N/P | √ (zero-shot) | N/P | √ (LoRA-SFT) | N/P | N/P |
| Long, et al (2024, Canada, USA)[28] | GPT-4 | N/P | √ (instruction-based prompting) | √ | N/P | N/P | N/P |
| Mashatian,et al (2024, USA)[29] | GPT-4 | N/P | √ (zero-shot, few-shot) | √ | N/P | N/P | N/P |
| Miao,et al (2024, USA)[30] | GPT-4 | N/P | √ (CoT technique) | √ | N/P | N/P | N/P |
| Murugan,et al (2024, USA)[31] | GPT-4 | N/P | √ | √ | N/P | N/P | N/P |
| Peng, et al (2023, USA)[32] | GPT-3 | N/P | √ (prompt tuning, zero-shot) | N/P | √ | N/P | N/P |
| Rau, et al (2023, Germany)[33] | GPT-3.5-turbo | N/P | √ (instruction-base d prompting) | N/P | N/P | √ | N/P |
| Singer, et al (2023, USA)[34] | GPT-4 | N/P | N/P | √ (LangChain technology) | N/P | N/P | N/P |
| Tan, et al (2024, China)[35] | Baichuan-7B | √ | N/P | N/P | √ (instruction fine-tuning: full-parameter fine-tuning) | N/P | N/P |
| Wu,et al (2024, China)[36] | LLaMA | N/P | √ (prompt tuning) | N/P | √ (instruction tuning) | N/P | N/P |
| Xu, et al (2024, China)[37] | ChatGLM-6B | N/P | N/P | √ (LangChain technology) | √ (LoRA & Freeze) | N/P | N/P |
| Yang, et al (2022, USA)[38] | BERT | √ | N/P | N/P | √(SFT) | N/P | N/P |
| Yu, et al (2024, UK)[39] | DialoGPT, ChatGPT-3.5 | N/P | √ | N/P | √ | N/P | N/P |
| Zheng, et al (2024, China)[40] | ChatGLM2-6B | N/P | √ (N/P, prompt tuning) | N/P | N/P | N/P | N/P |
| Zhu, et al (2024, USA)[41] | ChatGPT-4 | N/P | N/P | N/P | √ | N/P | N/P |
| Choi, et al (2024, USA)[42] | GPT-4 | N/P | N/P | √ (with multi-agent orchestration system) | N/P | N/P | N/P |
| Fu, et al (2024, China)[43] | RoBERT a | N/P | √ (template-based prompting, prompt tuning) | N/P | √(LoRA) | N/P | N/P |
| Jia, et al (2024, USA)[44] | The Gemma series, LLaMA collections, and the Mistral series | √ | N/P | N/P | √ (instruction fine-tuning) | N/P | N/P |
| Zhao, et al (2024, China)[45] | Aquila-7B | √ | N/P | N/P | √ (SFT, DPO-RLHF) | N/P | N/P |
| Li, et al (2024, USA)[46] | Mistral 7B | √ | N/P | N/P | √ (instruction tuning, LoRA) | N/P | N/P |
| Lammert, et al (2024, Germany)[47] | Gemini Pro | N/P | √ (CoT technology) | √ | N/P | N/P | N/P |
| Yang, et al (2024, China)[48] | Baichuan2-Base | √ | N/P | N/P | √ (Full-parameter SFT, LoRA-SFT) | N/P | N/P |
| Li, et al (2024, USA)[49] | GPT-3.5-turbo | N/P | √ (CoT technology) | √ (KG) | N/P | N/P | N/P |
| Na, et al (2024, Australia)[50] | GPT-3.5-turbo-16k | N/P | √ (template-based prompting) | N/P | √ (LoRA-SFT, instruction tuning) | N/P | N/P |
| Jia, et al (2024, China)[51] | LLaMA-7B | N/P | N/P | N/P | √ (LoRA) | N/P | N/P |
| Bhatti, et al (2023, Canada)[52] | Llama 2 70B | N/P | √ (template-based prompting) | N/P | √ (QLoRA-PEFT) | N/P | N/P |
| Chen, et al (2023, China)[53] | Baichuan2-7B-Base and Baichuan2-13B-Base models | √ | √ (template-based prompting) | N/P | √ (SFT, instruction tuning) | N/P | N/P |
| Ge, et al (2023, USA)[54] | Versa (GPT series) | N/P | N/P | √ | N/P | N/P | N/P |
| Liu, et al (2023, USA)[55] | LLaMA2 | N/P | N/P | N/P | √ (LoRA, instruction tuning) | N/P | N/P |
| Yang, et al (2023, China)[56] | Ziya-LLaMA-13B-v1 | √ | N/P | N/P | √ (SFT, RLHF, LoRA) | N/P | N/P |
| Haghighi, et al (2024, USA)[57] | LLaMA 2-7b-chat | √ | N/P | N/P | √ (QLoRA) | N/P | N/P |
| Kharitonova, et al (2024, Spain)[58] | GPT-3, LLaMA-1, LLaMA-2 | N/P | √ (template-based prompting) | √ | N/P | N/P | N/P |
| Labrak, et al (2024, France)[59] | Mistral 7B Instruct v0.1 | √ | √ (instruction-base dprompting, few-shot) | N/P | √ (QLoRA-SFT) | N/P | N/P |
| Lee, et al (2024, Korea)[60] | GPT 3.5 | N/P | √ (instruction-base d prompting) | √ (LangChain technology) | N/P | N/P | N/P |
| Li, et al (2023, USA)[61] | LLaMA-7B | N/P | √ (instruction-base d prompting) | N/P | √ | N/P | N/P |
| Liu, et al (2024, USA)[62] | Alpaca-7B | N/P | N/P | N/P | √ (instruction tuning, LoRA) | N/P | N/P |
| Tian, et al (2024, China)[63] | Ziya-13B-v2 | √ | √ (zero-shot, few-shot) | N/P | √ (SFT, RLHF) | N/P | N/P |
| Wang, et al (2023, China)[64] | LLaMa-7B | N/P | N/P | N/P | √ (instruction tuning) | N/P | N/P |
| Wang, et al (2023, China)[65] | LLaMA-33B | N/P | N/P | N/P | √ (QLoRA-SFT, RLHF) | N/P | N/P |
| Kim (2023, Korea)[66] | Llama-2-13B | N/P | N/P | N/P | √ | N/P | N/P |
| Kumichev, et al (2024, Russia)[67] | LLaMA-7b | N/P | √ (instruction-based prompting) | N/P | √ (LoRA) | N/P | N/P |
| Kweon, et al (2024, Korea)[68] | LLaMA | √ | √ (prompt tuning) | N/P | √ (instruction tuning) | N/P | N/P |
| Luo, et al (2024, China)[69] | Baichuan-13B | N/P | N/P | √ | √ (LoRA-SFT) | N/P | N/P |
| Shi, et al (2024, USA)[70] | Vicuna-7B | N/P | N/P | √ | N/P | √ | N/P |
| Yang, et al (2024, USA)[71] | GPT-3.5-turbo | N/P | √ (zero-shot, CoT technology) | N/P | N/P | N/P | N/P |
| Lozano, et al (2023, USA)[72] | GPT-3.5 and GPT-4 | N/P | N/P | √ (Langchain technology) | N/P | N/P | N/P |
| Yu, et al (2024, USA)[73] | GPT-4 | N/P | √ (few-shot) | N/P | N/P | √ (Langchain technology) | N/P |
| Zhang, et al (2023, China)[74] | ChatGPT | N/P | √ (instruction-based prompting) | N/P | √ (SFT, RLAIF) | N/P | N/P |

Information not provided is marked by N/P.

LoRA: Low-Rank Adaptation; RAG: Retrieval-augmented generation; KG: Knowledge Graph; SFT: Supervised Fine-tuning; RLHF: Reinforcement learning from human feedback; COT: Chain-of-Thought; QLoRA: Quantized Low-Rank Adaptation; RLAIF: reinforced learning from AI feedback; PEFT: Parameter-Efficient Fine-Tuning; DPO: Direct Preference Optimization; IT5: Text-to-Text Transfer Transformer for Italian; GPT: Generative Pretrained Transformer; LLaMA: Long-Language Model Anthropic; GLM: General Language Model; BERT: Bidirectional Encoder Representations from Transformers; RoBERT: Robustly Optimized BERT Pretraining Approach

Table 3 Evaluation metrics for process and usability assessment of LLMs in healthcare

| **Evaluation metrics domain** | **Evaluation metric** | **Associated study** | **Number** |
|---|---|---|---|
| Process evaluation metrics | Repetition testing turn | 33,34,45,53,56,63,74 | 7 |
| | Model size | 32,38,44,46,53,68,69 | 7 |
| | Used training parameters | 38,46,56,57,58 | 5 |
| | Response time | 17,20,21,33,46 | 5 |
| | Output length | 21,39,65 | 3 |
| | Response rate | 16,17 | 2 |
| | Likelihood of generating harmful content | 16 | 1 |
| | Average number of dialogues turns per item | 17 | 1 |
| | Percentage of free-text interaction usage | 17 | 1 |
| | Average number of tokens per second | 21 | 1 |
| Usability evaluation metrics | User helpfulness | 16,25,27,39,48,50,62,64,73 | 9 |
| | User intent | 16 | 1 |
| | Response costs fulfillment | 33 | 1 |
| | Personalization | 71 | 1 |
| | Interactivity | 71 | 1 |

Table 4 Evaluation metrics for model-based outcome of LLMs in healthcare

| Metrics Types | Evaluation metric | Associated study | Number |
|---|---|---|---|
| General domain standard metrics | Accuracy/correctness | 15,16,18,19,20,21,29,32,38,44,49,50,52,57,58,63,68,70,71,74 | 20 |
| | F1 score | 15,17,22,27,29,32,38,42,43,46,51,57,61,67,72 | 15 |
| | Precision | 17,22,27,29,32,38,43,46,51,57,61,67,72 | 13 |
| | Recall | 17,22,27,29,32,38,43,46,51,57,61,67,72 | 13 |
| | ROUGE | 24,25,35,43,46,48,55,63,69,72,74 | 11 |
| | BLEU | 24,35,39,43,46,48,50,63,69,74 | 10 |
| | Distinct | 25,43,48,74 | 4 |
| | METEOR score | 24,50,72 | 3 |
| | GLEU | 35,48,74 | 3 |
| | Cosine similarity | 47,50,65 | 3 |
| | Perplexity | 25,39 | 2 |
| | CHRF | 50,72 | 2 |
| | Exact Match score | 38,46 | 2 |
| | Specificity | 29 | 1 |
| | AUC with confidence | 42 | 1 |
| | Sentence-BERT embeddings | 69 | 1 |
| | Expected Calibration Error | 58 | 1 |
| Automated metrics evaluation via structured prompts in LLMs | Fluency/smoothness | 45,48,53,56,72 | 5 |
| | Relevance | 45,53,72 | 3 |
| | Accuracy | 48,53 | 2 |
| | Completeness | 45,53 | 2 |
| | Proficiency | 45,53 | 2 |
| | Consistency | 48.72 | 2 |
| | Safety | 56 | 1 |
| | Professionalism | 56 | 1 |
| | Coherence | 72 | 1 |

BERT: Bidirectional Encoder Representations from Transformers; BLEU: Bilingual Evaluation Understudy; ROUGE: Recall-Oriented Understudy for Gisting Evaluation; METEOR: Metric for Evaluation of Translation with Explicit Ordering; GLEU: Generalized Language Evaluation Understudy; AUC: Area Under the Curve; CHRF: Character F-score

Table 5 The expert-assessed outcomes and strategies for the outcome evaluation metrics of LLMs in healthcare

| Type | Evaluation metric | Definition | Related concepts | Evaluation strategies |
|---|---|---|---|---|
| Information Quality | Accuracy | The degree to which human evaluators correctly assess the model's responses in line with the intended task or knowledge. | Accuracy[14,17,20,21,22,24,26,27,31,33,35,36,37,40,41,53,54,59,68,74]<br>Correctness[15,33,34,42,67,73]<br>Inaccuracy[16,21]<br>Missing content[16,68,69]<br>Error[17,19]<br>Validity[28]<br>Factuality[29,48]<br>Precision[63] | (1) Comparison with human expert-provided answers used as the gold standard [14,15,16,17,21,22,24,25,28,29,33,34,36,67,68];<br>(2) Likert scale/Binary or Multiple-Choice evaluation scale [17,19,20,26,27,31,37,40,41,59,63,69,73];<br>(3) Unstructured feedback [34,35,42,53,54,66];<br>(4) Comparison of performance rankings with different models [48,74] |
| | Completeness | The extent to which the model provides all necessary and relevant information for a given task. | Completeness[15,17,37,41,63]<br>Thoroughness[20]<br>Comprehensiveness[26] | Likert scale[15,17,20,26,37,41,63] |
| | Relevance | How well the model's output aligns with the user's query or task requirements. | Relevance[16,20,25,26,31,37,42,50,62,70]<br>Clinical relevance[32]<br>Scientific consensus[69] | (1) Comparison with human expert-provided answers used as the gold standard[16];<br>(2) Likert scale/Binary or Multiple-Choice evaluation scale[20,25,26,31,32,37,42,50,62,69];<br>(3) Comparison with a set of predicted and actual responses[25];<br>(4) Unstructured feedback[70] |
| | Comprehension | The model's ability to understand and interpret the meaning and context of the input correctly. | Comprehension[16,17,29]<br>Contextual Understanding[36]<br>Comprehensibility[59]<br>Understandability[62,66,69]<br>Explainability[71] | (1) Comparison with human expert-provided answers used as the gold standard[16,29];<br>(2) Likert scale/Binary or Multiple-Choice evaluation scale [17,59,62,69];<br>(3) Comparison of performance rankings with different models[36]<br>(4) Unstructured feedback[66,71] |
| | Consistency | The model's ability to provide reliable, predictable, and consistent responses over time, maintaining performance across different interactions. | Stability[17]<br>Reliability[19]<br>Concordance[28,47]<br>Consistency[32-34] | (1) Likert scale[17,25,32];<br>(2) Unstructured feedback[19,47];<br>(3) Comparison with human expert-provided answers used as the gold standard[28,33,34] |
| Safety and Risk | Safety | The model's ability to prevent misuse or exploitation, ensuring safe interactions and protecting against malicious inputs. | Security[17]<br>Safety[28,48,64] | (1) Likert scale[17,64];<br>(2) Comparison with human expert-provided answers used as the gold standard[28]<br>(3) Comparison of performance rankings with different models[48] |
| | Risk | The potential negative outcomes or harm that could arise from a model's response or action. | Risk of causing severe harm[16]<br>Harm likelihood and harm extent[19]<br>Possible harm[29,69]<br>Risk[31,35,39,] | (1) Categories (Presence or absence)[16];<br>(2) Likert scale/Binary or Multiple-Choice evaluation scale[19,29,69];<br>(3) Comparison with human expert-provided answers used as the gold standard[29];<br>(4) Unstructured feedback[35,39]; |
| | Bias | The likelihood that the model's output reflects unfair, prejudiced, or skewed perspectives. | Bias[16,19,29,50,63,69] | (1) Categories (Presence or absence)[19];<br>(2) Likert scale/Binary or Multiple-Choice evaluation scale[16,63,69];<br>(3) Comparison with human expert-provided answers used as the gold standard[29];<br>(4) Unstructured feedback[50] |
| | Hallucination | The generation of false, inaccurate, or fabricated information by a model. | Hallucination[14,16,24,31,47] | (1) Categories (Presence or absence)[14,16,24];<br>(2) Comparison with human expert-provided answers used as the gold standard[47];<br>(3) Likert scale[31] |
| Reasoning and Justification | Provision of rationales with citations | The inclusion of logical explanations and references to credible sources in the model's output to support its answers. | Rationales and citations[18,31,47]<br>Reference integration[26]<br>Knowledge Correlation[36] | (2) Categories (Presence or absence)[18,47];<br>(3) Likert scale/Binary or Multiple-Choice evaluation scale[26,31,69];<br>(4) Performance ranking rates across models[36] |
| | Reasoning | The model's ability to logically analyze, infer, and derive conclusions or decisions based on available data or input. | Logic[25,35]<br>Reasoning[29,42]<br>Correct reasoning[16,69] | (1) Comparison with a set of predicted and actual responses[25];<br>(2) Comparison with human expert-provided answers used as the gold standard[16,29]<br>(3) Likert scale/Binary or Multiple-Choice evaluation scale[69];<br>(3) Unstructured feedback[35,42] |
| Communication Quality | Quality | A measure of how effectively the model addresses the user's query with expertise. | Capability[22]<br>Quality[25,30,35,39,45,53]<br>Competency[28]<br>Professionalism[48]<br>Proficiency[53] | (1) Unstructured feedback[22,30,35,39];<br>(2) Likert scale[25];<br>(3) Comparison with human expert-provided answers used as the gold standard[28]<br>(4) Comparison with a set of predicted and actual responses[45];<br>(5) Comparison of performance rankings with different models[48,53] |
| | Empathy | The evaluation of a response's suitability in a clinical context, assessing its professionalism, appropriateness, and sensitivity to patients. | Inappropriateness[19]<br>Empathy[27,50]<br>Sensitivity[35] | (1) Likert scale[19,27];<br>(2) Unstructured feedback[35,50] |
| | Readability | How easily the model's generated responses can be understood, focusing on vocabulary, sentence structure, clarity, and fluency | Coherence[20,62,73]<br>Fluency [22,25,36,63]<br>Language[31]<br>Readability[32,36,59,73]<br>Clarity[31]<br>Structure[50]<br>Conciseness[62]<br>Smoothness[64] | (1) Likert scale [20,25,31,32,37,41,50,59,62,63,64,73];<br>(2) Comparison with the gold standard provided by human experts[22];<br>(3) Comparison with a set of predicted and actual responses[25];<br>(4) Comparison of performance rankings with different models[36] |
| | Responsiveness | The ability to quickly and effectively react to changes, needs, or feedback. | Responsiveness[25,27,35]<br>Timeliness[37] | (1) Unstructured feedback[25,35];<br>(2) Likert scale[27,37] |
| | Creativity | The ability of the model to generate novel and original content. | Creativity[22] | (1) Comparison with human expert-provided answers used as the gold standard[22] |
| Cultural fidelity | Cultural fidelity | The faithful transmission and preservation of cultural traditions, values, and practices, ensuring their authenticity and completeness across generations. | Fidelity to cultural heritage[22] | (1) Comparison with human expert-provided answers used as the gold standard[22] |

PRISMA 2020 flow diagram for new systematic reviews which included searches of databases and registers only

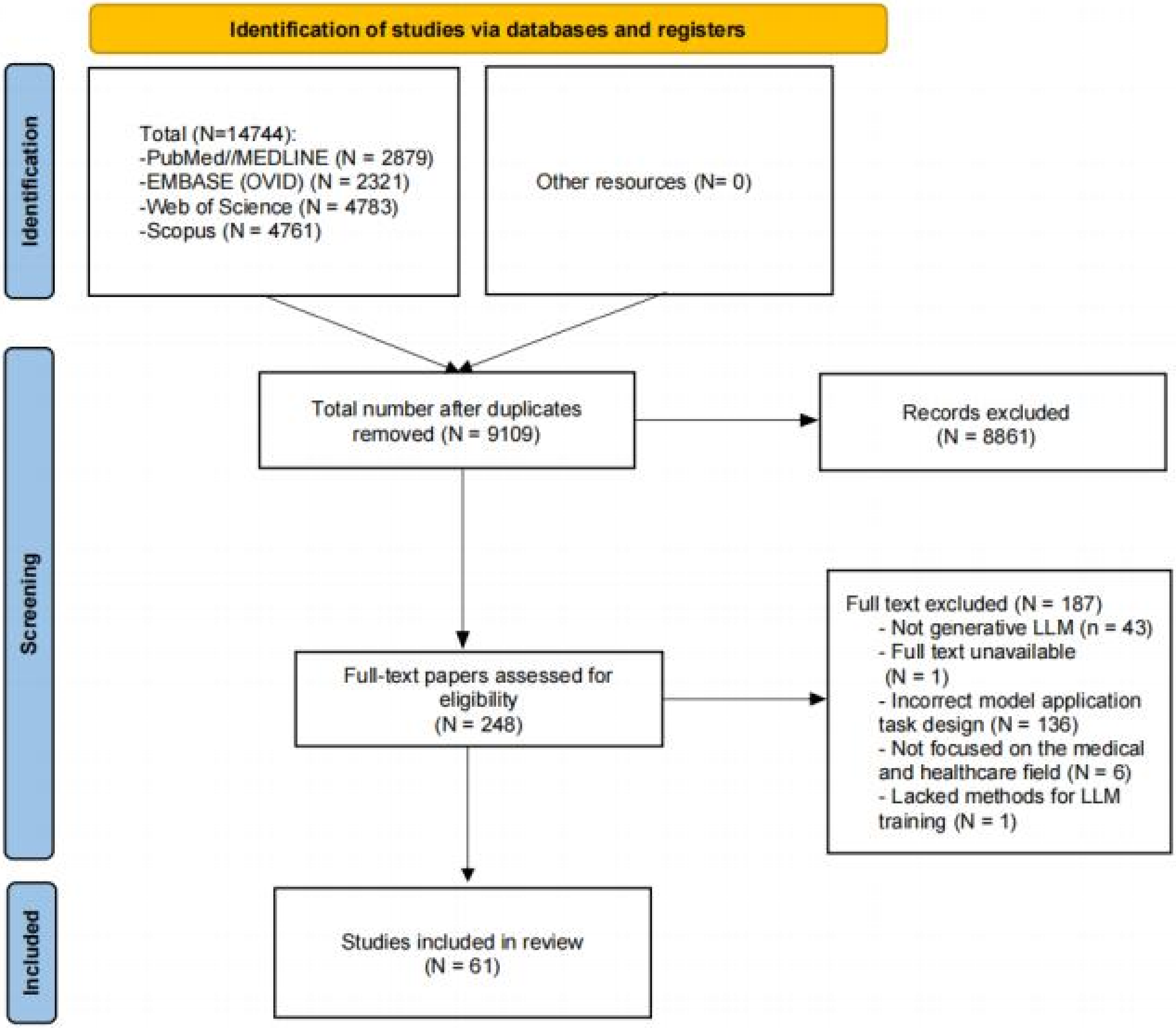